\pdfoutput=1

\documentclass[11pt]{article}
\usepackage{coling}
\usepackage{times}
\usepackage{latexsym}

\usepackage[T1]{fontenc}

\usepackage[utf8]{inputenc}
\usepackage{microtype}
\usepackage{inconsolata}

\usepackage{graphicx}
\usepackage{url}
\usepackage[misc]{ifsym}
\usepackage{latexsym}
\usepackage{epsfig}
\usepackage{float}
\usepackage{algorithm}
\usepackage{algorithmic}
\usepackage{enumitem}
\usepackage{tabularx} 
\usepackage{booktabs} 
\usepackage{epsfig}
\usepackage{epstopdf}
\usepackage{amsmath}
\usepackage{amssymb}
\usepackage{graphics}
\usepackage{graphicx}
\usepackage{epsfig}
\usepackage{subfigure}
\usepackage{subcaption}
\usepackage{fancyhdr}
\usepackage{array}
\usepackage{placeins} 
\usepackage{multirow}
\usepackage{eucal}
\usepackage{makecell}

\usepackage{color}
\usepackage{svg}

\newcommand{\pheadWithSpace}[1] {\vspace{2mm}\noindent\textbf{#1.}}

\begin{document}
\title{CEGRL-TKGR: A Causal Enhanced Graph Representation Learning Framework for Temporal Knowledge Graph Reasoning}
\author{
 \textbf{Jinze Sun\textsuperscript{1}},
 \textbf{Yongpan Sheng\textsuperscript{1,2}\textsuperscript{(\Letter)}},
 \textbf{Lirong He\textsuperscript{3}},
 \textbf{Yongbin Qin\textsuperscript{4}},
 \textbf{Ming Liu\textsuperscript{5}},
 \textbf{Tao Jia\textsuperscript{1}}
\\
\\
 \textsuperscript{1}College of Computer and Information Science, Southwest University, Chongqing 400715, China
  \\
 \textsuperscript{2}School of Big Data \& Software Engineering, Chongqing University, Chongqing 401331, China
 \\
 \textsuperscript{3}School of Information Science and Engineering, Chongqing Jiaotong University, Chongqing 400074, China
 \\
 \textsuperscript{4}College of Computer Science and Technology, The State Key Laboratory of Public Big Data, Guizhou University
 \\
 \textsuperscript{5}Deakin University, 3125, Victoria, Australia
\\
 \small{
   {ssunjinze@outlook.com \quad \{shengyp2011,ronghe1217\}@gmail.com \quad ybqin@gzu.edu.cn \quad m.liu@deakin.edu.au \quad tjia@swu.edu.cn} 
 }
}

\maketitle

\vspace{-20pt}

\begin{abstract}
Temporal knowledge graph reasoning (TKGR) is increasingly gaining attention for its ability to extrapolate new events from historical data, thereby enriching the inherently incomplete temporal knowledge graphs. Existing graph-based representation learning frameworks have made significant strides in developing evolving representations for both entities and relational embeddings. Despite these achievements, there's a notable tendency in these models to inadvertently learn biased data representations and mine spurious correlations, consequently failing to discern the causal relationships between events. This often leads to incorrect predictions based on these false correlations. To address this, we propose an innovative \textbf{C}ausal \textbf{E}nhanced \textbf{G}raph \textbf{R}epresentation \textbf{L}earning framework for TKGR (named CEGRL-TKGR). This framework introduces causal structures in graph-based representation learning to unveil the essential causal relationships between events, ultimately enhancing the performance of the TKGR task. Speciﬁcally, we first disentangle the evolutionary representations of entities and relations in a temporal knowledge graph sequence into two distinct components, namely causal representations and confounding representations. Then, drawing on causal intervention theory, we advocate the utilization of causal representations for predictions, aiming to mitigate the effects of erroneous correlations caused by confounding features, thus achieving more robust and accurate predictions. Finally, extensive experimental results on six benchmark datasets demonstrate the superior performance of our model in the link prediction task. 
\end{abstract}

\section{Introduction}
Knowledge graphs (KGs) have gained significant promise in natural language processing or knowledge engineering perception tasks~\cite{chen2022overview}. They model real-world factual knowledge using multi-relationship graph structures. However, factual knowledge in reality is constantly evolving, resulting in the form of event knowledge. This has led to the development and application of temporal knowledge graphs (TKGs). TKG encodes the relationship information of entities and events and their timing for capturing the dynamics of entities and their relationships over time~\cite{gastinger2022evaluation}. Thus, analyzing the TKG provides a comprehensive understanding of the evolving events, based on which various time-dependent applications have been developed, including time-sensitive semantic search~\cite{barbosa2013shallow}, policy making~\cite{deng2020dynamic}, stock forecasting~\cite{feng2019temporal}, and more~\cite{chen2022overview}.

The reliability of applications depends on accurate predicting, which highly relies on data integrality. However, existing TKGs are inevitably incomplete due to the partial observation of real-world~\cite{liang2022reasoning}. To address this limitation and enhance the representation capability of the TKG, temporal knowledge graph reasoning~(TKGR) models are proposed and aim to extrapolate new facts and relationships in the TKG according to their historical temporal information. Existing models explore different strategies to achieve satisfactory results on the TKGR task. GHNN~\cite{han2020graph} and GHT~\cite{sun2022graph} model historical facts as point-in-time processes. TKGR-RHETNE~\cite{sun2023tkgr} jointly models the relevant historical event and temporal neighborhood event context of events in the TKG. RE-NET~\cite{jin2020recurrent} and RE-GCN~\cite{li2021temporal} introduce graph neural networks~(GNN) into sequence models to capture structural and temporal dependencies between entities.
TKGR-GPRSCL~\cite{xiong2024tkgr} captures complex structure-aware information by encoding paths across entities and obtaining temporal correlations in the complex plane. TLogic~\cite{liu2022tlogic} and TITer~\cite{sun2021timetraveler} design interpretable models based on logical rules and reinforcement learning, respectively. Despite the achievements of previous studies, they have overlooked the reality that there are numerous confounding factors in the TKG, such as shallow patterns and noisy links. However, these confounding factors commonly misguide the reasoning process in the TKG, resulting in the acquisition of incorrect dependencies and the generation of non-causal predictions~\cite{sui2022causal}. 

To address the aforementioned issues, we advocate for the integration of causal theory into TKGR to guide learning of the essential causal relationships between events and mitigate the impact of confounding factors on the TKGR task. Specifically, we first construct a structural causal model~\cite{zevcevic2021relating} to comprehensively analyze and model the TKGR task from a causal perspective. Then, based on the causal model, we propose a new framework, namely \textbf{C}ausal \textbf{E}nhanced \textbf{G}raph \textbf{R}epresentation \textbf{L}earning (CEGRL-TKGR), to disentangle confounding factors from the essential causal factors in the TKG. \emph{To the best of our knowledge, this is the first study to incorporate causal intervention in a graph representation learning framework for learning the evolutionary representations of entities and relations in the TKG}.
To conclude, our contributions in this paper are 3-folds:
\begin{itemize}
\item We propose a novel \textbf{C}ausal \textbf{E}nhanced \textbf{G}raph \textbf{R}epresentation \textbf{L}earning framework for \textbf{T}emporal \textbf{K}nowledge \textbf{G}raph \textbf{R}easoning, called CEGRL-TKGR, to uncover the essential causal relationships between events and mitigate the impact of confounding factors. 

\item The proposed CEGRL-TKGR framework disentangles the evolutionary representations of entities and relations into causal and confounding representations. Then, it applies causal interventions to perform backdoor adjustments of representations, prioritizing predicted causal features while minimizing the impact of spurious correlations introduced by confounding features.

\item Comprehensive experimental results demonstrate that CEGRL-TKGR  outperforms state-of-the-art baselines on six real-world datasets in the link prediction task. Further, comprehensive studies confirm the contribution of the introduced causal structures and interventions\footnote{To illustrate the evaluation of our  CEGRL-TKGR framework and facilitate further research on this topic, we have made the experimental details and source code of the framework publicly available at \url{https://github.com/shengyp/CEGRL-TKGR}.}.
\end{itemize}

\section{Related Work}
\label{sec:related}
\subsection{Temporal Knowledge Graph Reasoning}
TKGR in extrapolation settings focuses on predicting new facts about the future based on historical events. Specifically, CyGNet~\cite{zhu2021learning} uses a copy-generating mechanism to capture the global repetition rate of facts. GHNN~\cite{han2020graph} and GHT~\cite{sun2022graph} construct a temporal point process (TPP) to capture the temporal dynamics of successive events, predicting future facts by estimating the conditional probability of the TPP. In recent years, with the successful application of GNN in many dynamic scenarios~\cite{zhang2022dynamic}, they have also been introduced into structural-semantic dependency models in TKGR. RE-NET~\cite{jin2020recurrent} used a neighborhood aggregator and cyclic event encoder to model historical facts as subgraph sequences. RE-GCN~\cite{li2021temporal} uses RGCN~\cite{schlichtkrull2018modeling} to learn evolutionary representations of entities and relationships at each timestamp. CEN~\cite{li2022complex} uses length-aware convolutional neural networks (CNNS) to process evolutionary patterns of different lengths. There are also some studies to solve the TKGR problem through path search. For example, TLogic~\cite{liu2022tlogic} completes link prediction tasks based on temporal logic rules learned from temporal knowledge graphs. TITer~\cite{sun2021timetraveler} proposes a TKG prediction model based on reinforcement learning, which uses time-shaped rewards based on Dirichlet distribution to guide model training. All of the methods discussed above have limitations in modeling entity and relationship representations, in particular ignoring cause-and-effect relationships between different entities, which we believe is key to making correct predictions.

\subsection{Causal Representation Learning}
In graph causal representation learning, researchers have explored various methods to improve the explanatory power and generalization performance of GNNs. By applying the principles of causal reasoning to graph-structured data, the researchers sought to address the challenges GNNs face when dealing with complex systems such as social networks, molecular maps, and syntax trees of program code. DIR~\cite{wu2021discovering} is proposed to reveal the intrinsic interpretability of GNNs by discovering invariant reasons, which involves splitting input graphs into causal and non-causal fruit graphs and training the two classifiers through invariant risk loss functions. GOOD~\cite{chen2022invariance} improves the cross-domain generalization of graphs by distinguishing invariant subgraphs from other parts of graphs that are susceptible to domain transfer. CAL~\cite{sui2022causal} introduces de-confounding training to distinguish the key and secondary parts of the graph and eliminate the confounding effect of the secondary parts on model prediction. CFLP~\cite{zhao2022learning} points out that the causal relationship between graph structure and link presence is often ignored, and proposed to generate counterfactual links to enhance training data and reduce reliance on false associations. Zevcevic~\emph{et al.}~\cite{zevcevic2021relating} theoretically analyze the relationship between GNNs and structural causal models (SCMs) and design a new class of neuro-causal models. However, none of the work has been done to combine causal learning with the TKGR task.

\section{Preliminary}
\subsection{Notations and Task Formulation}
A TKG $\mathcal{G}$ can be formalized as a sequence of knowledge graph slices $\left\{\mathcal{G}_{1}, \mathcal{G}_{2}, \ldots ., \mathcal{G}_{T}\right\}$, where $\mathcal{G}_{t} = \left\{\left(e_{s}, r, e_{o}, t\right) \in \mathcal{G}\right\}$ denotes a knowledge graph slice that consists of facts that occurred at the timestamp $t$ range from $t_{0}$ to $t_{n}$. Here, $e_{s}$ and $e_{o}$ represent the subject and object entities, respectively, and $r$ denotes the predicate as a relation type. Besides, $\mathbf{e}_{s}$, $\mathbf{r}$, $\mathbf{e}_{o}$ written in bold represent their embeddings. The objective of TKGR task is to predict 
either the subject in a give query $\left(?, r, e_{o}, t\right)$ or the object in a given query $\left(e_{s}, r, ?, t\right)$ with $t > t_{n}$.

\subsection{A Causal Perspective on the GNN-Based TKGR Task}
\subsubsection{GNN-based TKGR Paradigm}
Inspired by previous GNN-based modeling in a casual look~\cite{didelez2001causality,sui2022causal}, we abstract the GNN-based TKGR process through a structural causal figure, as shown in Fig.~\ref{fig:causal graph}, encompassing five distinct variables. The connectivity from one variable to another epitomizes the causal relationship, delineated as the cause $\rightarrow$ effect. The variables are described as follows:
\begin{figure}[t]
    \centering
\includegraphics[width=0.32\textwidth]{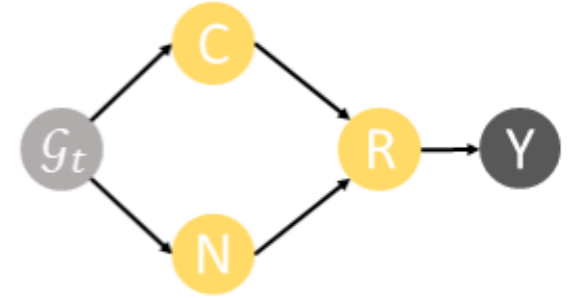}
	\caption{The GNN-based structural causal graph for the TKGR task.}
	\label{fig:causal graph}
\end{figure}

\begin{itemize}
    \item Graph data $\mathcal{G}_{t}$: The knowledge graph at each timestamp $t$, manifests as a directed multi-relationship figure.
    
    \item Causal Feature $C$: These features epitomize the causal essence of the targeted entity, providing a fundamental understanding of its inherent dynamics.
    
    \item Confounding Feature $N$: These features, discerned from GNN, embody the confounding attributes, unveiling the potential biases or trivial patterns ingrained in graph-based learning methodologies.
    
    \item Representation $R$: These representations are the entity and relational representations of the output of the final GNN layer after learning for $\mathcal{G}_{t}$.
    
    \item Prediction $Y$: Denoted as TKGR as the link prediction, this aspect transitions through the decoder, rendering the ultimate reasoning based on the preceding representation.
\end{itemize}

The causal embedding encapsulates the causal features $C$, authentically mirroring the implicit knowledge inherent in the knowledge graph $\mathcal{G}_{t}$. Conversely, $N$ symbolizes the confounding features, which might be spawned by data biases, data noise, or superficial patterns within graph-based learning methodologies. These confounding features forge a backdoor pathway between $C$ and $Y$, fostering spurious correlations that don’t contribute to accurate reasoning. Functionally, the structural operation denoted by $C$ $\rightarrow$ $R$ $\leftarrow$ $N$ portrays a GNN, wherein both the causal features $C$ and the confounding features $N$, as discerned by the target entity from the graph data, exert a direct impact on the output $R$ of the GNN. Subsequently, the output $R$ of GNN directly sways the model inference outcome, illustrated as $R$ $\rightarrow$ $Y$.

In the graph-based TKGR paradigm, causal and confounding features are not decoupled for each entity or relationship embedding. Using causal graphs, we aspire to explicitly separate causal embeddings and confounding embeddings from entity or relational representations, and aim to mitigate the effects of confounding features by performing causal interventions. This endeavor not only clarifies the inference process but also endeavors to refine the accuracy and reliability of the GNN-based TKGR mechanism.

\subsection{Causal Intervention Strategies}
Beyond fostering a novel comprehension of GNN-based TKGR, causal theory avails analytical instruments predicated on causal figures, such as causal intervention. Causal intervention facilitates a profound examination of the factors precipitating inference outcomes. As delineated by Fig.~\ref{fig:causal graph}., confounding feature $N$ and causal feature $C$ can be discerned from the knowledge graph $\mathcal{G}_{t}$. These features are contemplated in the representation $R$ of entities and relations, thereby establishing a backdoor pathway represented as $N$ $\leftarrow$ $\mathcal{G}_{t}$ $\rightarrow$ $C$ $\rightarrow$ $R$ $\rightarrow$ $Y$, with $N$ serving as the quick bridge between $C$ and $Y$.

To orchestrate a causal prognosis hinging on the causal feature C, it necessitates the modeling of $P(Y\mid C)$. However, the backdoor path distorts the probability distribution $P(Y\mid C)$ through the confounding effect of $N$, thereby necessitating the disentanglement of the backdoor pathway from $N$ to $Y$. It is imperative to stymie this backdoor pathway to mitigate the repercussions of the hybrid embedding, thereby enabling the model to reason robustly by leveraging the causal feature to the fullest. Causality theory is a potent toolkit to address this backdoor path dilemma.

We engage the do-calculus for executing causal interventions on variable $C$, intending to sever the backdoor path $N$ $\leftarrow$ $\mathcal{G}_{t}$ $\rightarrow$ $C$ $\rightarrow$ $R$ $\rightarrow$ Y. Our objective is to estimate $P(Y\mid do(C))$, as opposed to muddling it with $P(Y\mid C)$. By using Bayes' theorem with the causal postulation, we can extrapolate the ensuing expression:
\begin{align}
  P(Y\mid do(C)) = \sum_{n \in N} P(Y\mid C, n)P(n).
\label{eq:bda} 
\end{align}
The equation above illustrates that to gauge the causal influence of $C$ on $Y$, it's requisite to take into account the inference outcomes of both causal and confounding features. This can be perceived as re-coupling the disentanglement feature embeddings, utilizing them for deductive reasoning at future timestamps. However, $C$ and $N$ are usually unobservable, and it is difficult to obtain them directly at the data level, which makes the calculation of the Eq.~(\ref{eq:bda}) very challenging. In the next section, we discuss ways to overcome this problem.

\section{The Proposed CEGRL-TKGR Framework}
\label{sec:2}
\subsection{The Overall Architecture of TKGR-GPRSCL}
\begin{figure*}[t]
    \centering
\includegraphics[width=\textwidth]{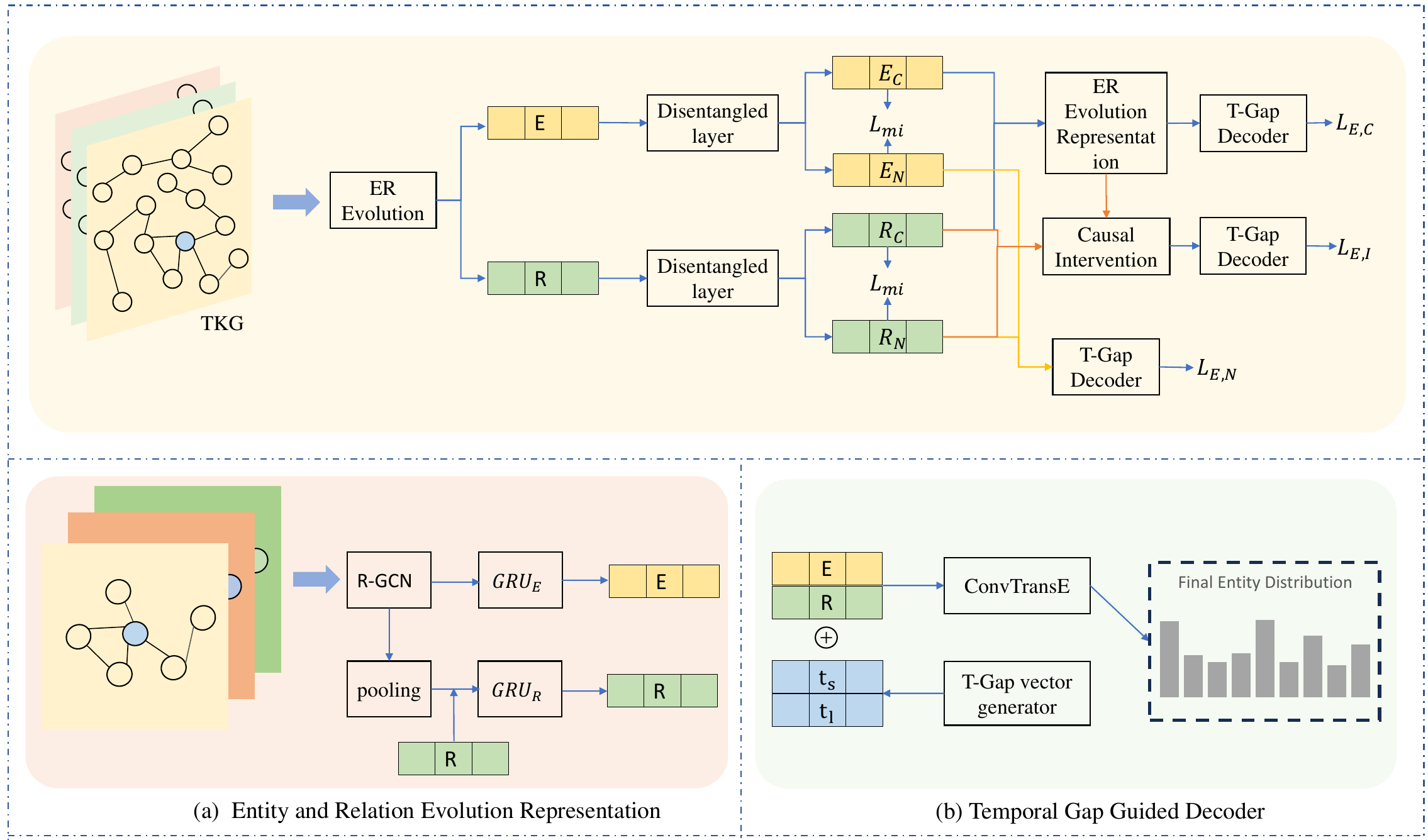}
	\caption{The overall architecture of our proposed CEGRL-TKGR framework.}
	\label{fig:archi}
 \vspace{-15pt}
\end{figure*}

We detail the CEGRL-TKGR framework for learning representations of entities and relationships based on causal features and confounding features. CEGRL-TKGR consists of three parts: (1) The representation learning part that learns the structure dependence in each $\mathcal{G}_{t}$; (2) The decoupling learning part that learns the entity and relation representations; (3) The decoder part that is designed based on the time interval. The overall architecture of the framework is shown in Fig.~\ref{fig:archi}.

\subsection{Entity and Relation Evolution Representation}
Within each $\mathcal{G}_{t}$, representation learning of entities and relationships involves the aggregation of multiple relationships, as well as information from multiple hop neighbors under a single timestamp. Between adjacent $\mathcal{G}_{t}$, we expect to accurately capture the order dependencies inherent in the subgraph with different timestamps. Drawing inspiration from the RE-GCN model~\cite{li2021temporal}, we employ the $\omega$-layer RGCN, which hinges on structure modeling and a recurrent mechanism to progressively update the representations of entities and relations. This approach allows for a more nuanced understanding and modeling of the dynamic interactions within the graph over time. 
\begin{align}
\mathbf{e}^{l+1}_{o,t} &= \operatorname{RReLu}\Bigg(
\sum_{(e_{s}, r, e_{o}) \in \mathcal{G}_{t}} \frac{1}{d_{e_{o}}} \mathbf{W}^{l}_{1}\left(\Phi\left(\mathbf{e}^{l}_{s,t}, \mathbf{r}_{t}\right)\right) \notag \\
&\qquad + \mathbf{W}^{l}_{2} \mathbf{e}^{l}_{o,t} \Bigg),
\label{eq:RGCN} \\
\mathbf{E}_{t} &= \operatorname{GRU}\left(\mathbf{E}_{t-1}, \mathbf{E}^{\prime}_{t}\right).
\label{eq:GRU}
\end{align}

In the Eq.~(\ref{eq:RGCN}), we describe how the embedding $\mathbf{e}^{l+1}_{o,t}$ of entity $e_{o}$ at time step $t$ and layer $l+1$ is computed. We integrate the information of all entities and relations connected to entity $e_{o}$ in the knowledge graph $\mathcal{G}_{t}$. $\mathbf{W}^{l}_{1}, \mathbf{W}^{l}_{2}$ is learnable weights and $\Phi$ has the option of addition or one-dimensional convolution. In the Eq.~(\ref{eq:GRU}), we showcase how the entity embedding matrix $\mathbf{E}_{t}$ is updated via the GRU. Specifically, we take the entity embedding matrix $\mathbf{E}_{t-1}$ at the previous time step $t-1$ and the aggregated entity embedding matrix $\mathbf{E}^{\prime}_{t}$ as inputs to obtain the entity embedding matrix $ \mathbf{E}_{t}$ at the current time step $t$.

For relations, ensuring consistency with the entity embedding updates within the subgraph sequence is crucial. To achieve this consistency, a specialized GRU tailored for relations is employed for the update process. This mechanism facilitates a harmonized evolution of both entity and relation causal embeddings over the sequence of subgraphs:
\begin{align}
\mathbf{r}_{t}^
{\prime}  &=\operatorname{pooling}\left(\mathbf{E}_{t-1}, R_{t}\right) \oplus \mathbf{r}, \\
\mathbf{R}_{t} &=\operatorname{GRU}\left(\mathbf{R}_{t-1}, \mathbf{R}^{\prime}_{t}\right),
\end{align} where $\mathbf{r}^{\prime}_{t}$ is an aggregation of all entities connected to relation $r$ via a mean pooling operation, and $\mathbf{R}^{\prime}_{t}$ is obtained by concatenating this result with the embeddings of all relations. Eventually, we update the relation embedding matrix $\mathbf{R}_{t}$ using a GRU.

\subsection{Disentangled Causal and Confounding Features}
In the previous subsection, the entity and relation representations are learned based on GNN-contained causal and confounding factors, and we separate them at the presentation level, which provides a solution to the previously mentioned problem of not being able to separate these two features at the data level. To do this, we introduce a decoupling module to decouple causal and confounding features. Taking the entity embedding matrix as an example, it is represented as follows:
\begin{align}
     \textbf{D}_{E,C}, \textbf{D}_{E,N} = \operatorname{softmax}(\operatorname{MLP}(\textbf{E})),   \\
         \textbf{E}_{C} = \textbf{E} \odot \textbf{D}_{E,C}, \textbf{E}_{N} = \textbf{E} \odot \textbf{D}_{E,N}.
\end{align}

We want the two embeddings learned from the decoupling module to be as independent as possible, which is essential to accurately separate causal and confounding features~\cite{chen2023causality}. Mutual information is a basic quantity to measure the nonlinear correlation of two random variables. Minimizing mutual information is a feasible scheme to decouple causal features from confounding features. Specifically, we implement this process with contrastive log-ratio upper-bound MI estimator~\cite{cheng2020club,wu2021disenkgat}, which utilizes variational distributions $q$ and a neural network to approximate the true distribution. We define the objective function as follows:
\begin{equation}
\begin{split}
\mathcal{L}_{mi} = &\ \mathbb{E}_{p(\textbf{E}_{C}, \textbf{E}_{N})} \left[ \log q_{\theta}(\textbf{E}_{N} | \textbf{E}_{C}) \right] \\
& - \mathbb{E}_{p(\textbf{E}_{C})}\mathbb{E}_{p(\textbf{E}_N)} \left[ \log q_{\theta}(\textbf{E}_{N} | \textbf{E}^{\prime}_{C}) \right].
\end{split}
\end{equation}

We perform the same operation with relation embedding decoupling, after which we obtain $\textbf{R}_{C}$ and $\textbf{R}_{N}$. 

\subsection{Temporal Gap Guided Decoder}
After the causal and confounding embeddings of entities and relations in the derived data, we use a specially crafted decoder to determine the likelihood score of potential entities and relations. Events or facts in a data stream may span different periods. For example, major political events may occur in rapid succession over a short period, while certain rare natural phenomena may occur sporadically and at longer intervals. With this in mind, it is reasonable to consider the time intervals of events to get an accurate picture of their temporal relationship. The key to the design of our decoder is the time interval vector, which guides the decoding process in considering the event time interval while calculating the fraction. Formulaic as:
\begin{equation}
\mathbf{t}_{s} =\boldsymbol{\alpha}_{s} t+\boldsymbol{\beta}_{s}, \ 
\mathbf{t}_{l} =\boldsymbol{\alpha}_{l} t+\boldsymbol{\beta}_{l}.
\end{equation}

Here, $\boldsymbol{\alpha}_{s}, \boldsymbol{\beta}_{s}, \boldsymbol{\alpha}_{l}$, and $\boldsymbol{\beta}_{l}$ signify learnable parameters. Adopting ConvTransE as our decoder, we introduce four variables, which traverse a one-dimensional convolutional layer followed by a fully connected layer, culminating in the extraction of a probability vector encompassing all entities. This process is mathematically articulated as:
\begin{equation}
\begin{split}
\mathbf{p}_{C}\left(e_{o} \mid e_{s},r,t\right) = &\ \operatorname{ReLU}\Big(\operatorname{ConvTransE}\big(\mathbf{e}_{s,C,t},  \\
& \mathbf{r}_{C,t}, \mathbf{t}_{s}, \mathbf{t}_{l}\big)\Big)\mathbf{E}_{C,t}.
\end{split}
\end{equation}

We apply the same decoding process to the confounding features to get $\mathbf{p}_{N}\left(e_{o} \mid e_{s},r,t\right)$.
\subsection{Causal Intervention and Training Objective}
Causal-based embedding learns the intrinsic causes that cause events to occur, so the reasoning results obtained from causal-based embedding are expected to yield reasonable input results. We define the supervised classification loss as follows:
\begin{equation}
\mathcal{L}_{E,C} = \sum_{(e_{s}, r, e_{o}, t) \in \mathcal{G}} \textbf{y}_{t}\log \mathbf{p}_{C}\left(e_{o} \mid e_{s},r,t\right),
\end{equation} where $\textbf{y}_{t}$ is label vector. Conversely, confounding features are conceptualized to address conceivable biases or superficial patterns emanating from the training dataset. Given their inability to aid in inference, we proceed to compute their output average across all entity categories and encapsulate the loss as:
\begin{equation}
\begin{split}
\mathcal{L}_{E,N} = &\ \frac{1}{\lvert \textbf{E}_{N,t} \rvert} \sum_{(e_{s},r,e_{o},t) \in \mathcal{G}} \operatorname{KL}\Big(\textbf{y}_{u}, \\
&\ \log\mathbf{p}_{N}\left(e_{o} \mid e_{s},r,t\right)\Big),
\end{split}
\end{equation} where KL denotes the KL-Divergence, $\textbf{y}_{u}$ represents the uniform distribution.

We believe that causal intervention is the manifestation of causal features under the influence of confounding features, but we cannot directly conduct causal intervention at the data level to mitigate confounding effects. Therefore, we obtain intervention features that combine causal features and confounding features at the representation level of entities and relationships. Specifically, according to the backdoor adjustment Eq.~(\ref{eq:bda}), we first introduce a random addition procedure to obtain the intervention feature, and for the intervention feature we expect the decoder to still output the correct result:
\begin{align}
    \mathbf{E}_{I,t} &= \phi\left(\mathbf{E}_{C,t},\mathbf{E}_{N,t}^{\prime}\right), \\
    \mathbf{p}_{I}\left(e_{o} \mid e_{s},r,t\right) &= \operatorname{ReLU}\Big(\operatorname{ConvTransE}\big(\mathbf{e}_{s,I,t}, \mathbf{r}_{I,t}, \notag \\
    &\qquad \mathbf{t}_{s}, \mathbf{t}_{l}\big)\Big)\mathbf{E}_{I,t},
\end{align}

where $\mathbf{E}_{N,t}^{\prime}$ is the confounding feature of the entites randomly sampled from $\mathbf{E}_{N,t}$. Then we define the loss as follows:
\begin{equation}
\mathcal{L}_{E,I} = \sum_{(e_{s}, r, e_{o}, t) \in \mathcal{G}} \textbf{y}_{t}\log \mathbf{p}_{I}\left(e_{o} \mid e_{s},r,t\right).
\end{equation}

Finally, the loss function of the model for the link prediction task is as follows:
\begin{equation}
\mathcal{L}_{E} = \mathcal{L}_{E,C} + \lambda_{1}\mathcal{L}_{E,N}+\lambda_{2}\mathcal{L}_{mi}+\lambda_{3}\mathcal{L}_{E,I},
\end{equation} where $\lambda_{1},\lambda_{2},\lambda_{3}$ are designated as hyper-parameters, and the first two are used to determine the strength of decoupled learning of the model and the latter is used to determine the strength of causal intervention of the model.

\section{Experiments and Analysis}
\label{sec:exp-and-analy}
\subsection{Experimental Settings and Implementation Details}
\noindent\textbf{Datasets}. We evaluate our model and baselines on six benchmark datasets, including ICEWS14~\cite{garcia2018learning}, ICEWS18~\cite{jin2019recurrent}, ICEWS05-15~\cite{garcia2018learning}, YAGO~\cite{mahdisoltani2014yago3}, WIKI~\cite{leblay2018deriving} and GDELT~\cite{leetaru2013gdelt}.  Statistics of the datasets are summarized in Table~\ref{tab:datasets}.

\begin{table}[htbp]
\centering
\caption{Statistics of datasets in the experiments.}
\label{tab:datasets}\scalebox{0.53}{
\begin{tabular}{@{}ccccccc@{}}
\toprule
Dataset   & \# Entity & \# Predict & \# Train & \# Valid & \# Test & Time interval \\ \midrule
ICEWS14   & 7128      & 230        & 63685    & 13823    & 13222   & 24 hours      \\
ICEWS18   & 23033     & 256        & 373018   & 45995    & 49545   & 24 hours      \\
ICEWS05-15 & 10488     & 251        & 322958   & 69224    & 69147   & 24 hours      \\
YAGO      & 10623     & 10         & 161540   & 19523    & 20026   & 1 year        \\
WIKI      & 12554     & 24         & 539286   & 67583    & 63110   & 1 year        \\ 
 GDELT& 7691& 240& 1734399& 238765& 305241&15 mins\\
\bottomrule
\end{tabular}}
\end{table}

\begin{table*}[t]
\centering
\caption{Experimental results of link prediction on ICEWS series dataset. The best result in each column is boldfaced. }
\label{tab:icewsres}\scalebox{0.61}{
\begin{tabular}{ccccccccccccc}
\hline
Model       & \multicolumn{4}{c}{ICEWS14}       & \multicolumn{4}{c}{ICEWS18}       & \multicolumn{4}{c}{ICEWS05-15}    \\ \hline
            & MRR   & Hits@1 & Hits@3 & Hits@10 & MRR   & Hits@1 & Hits@3 & Hits@10 & MRR   & Hits@1 & Hits@3 & Hits@10 \\
TransE      & 22.48 & 13.36  & 25.63  & 41.23   & 12.24 & 5.84   & 12.81  & 25.10   & 22.55 & 13.05  & 25.61  & 42.05   \\
Distmult    & 27.67 & 18.16  & 31.15  & 46.96   & 10.17 & 4.52   & 10.33  & 21.25   & 28.73 & 19.33  & 32.19  & 47.54   \\
ComplEx     & 30.84 & 21.51  & 34.48  & 49.59   & 21.01 & 11.87  & 23.47  & 39.97   & 31.69 & 21.44  & 35.74  & 52.04   \\
R-GCN       & 28.03 & 19.42  & 31.95  & 44.833  & 15.05 & 8.31   & 16.49  & 29.00   & 27.13 & 18.83  & 30.41  & 43.16   \\ \hline
TTransE     & 13.43 & 3.11   & 17.32  & 34.55   & 8.31  & 1.92   & 8.56   & 21.89   & 15.71 & 5.00   & 19.72  & 38.02   \\
TA-DistMult & 26.47 & 17.09  & 30.22  & 45.41   & 16.75 & 8.61   & 18.41  & 33.59   & 24.31 & 14.58  & 27.92  & 44.21   \\
TNTComplEx  & 32.12 & 23.35  & 36.03  & 49.13   & 21.23 & 13.28  & 24.02  & 36.91   & 27.54 & 19.52  & 30.80  & 42.86   \\
Evo-KG      & 26.90 & 16.69  & 30.57  & 47.39   & 25.46 & 16.25  & 29.15  & 43.21   & 26.32 & 15.82  & 31.96  & 50.80   \\
xERTE       & 40.79 & 32.70  & 45.67  & 57.30   & 29.31 & 21.03  & 33.51  & 46.48   & 46.62 & 37.84  & 52.31  & 63.92   \\
TITer       & 40.59 & 31.41  & 45.47  & 57.62   & 29.55 & 21.37  & 33.10  & 44.87   & 46.62 & 36.46  & 52.29  & 65.23   \\
TLogic      & 41.80 & 31.93  & 47.23  & 60.53   & 28.41 & 18.74  & 32.71  & 47.97   & 45.99 & 34.49  & 52.89  & 67.39   \\
RE-GCN      & 42.00 & 31.63  & 47.20  & 61.65   & 32.62 & 22.39  & 36.79  & 52.68   & 48.03 & 37.33  & 53.90  & \textbf{68.51}   \\
CEN         & 41.93 & 31.71  & 46.86  & 61.36   & 29.41 & 19.60  & 33.91  & 49.97   & 47.04 & 36.58  & 52.60  & 67.18   \\
GHT         & 38.28& 28.43& 42.85& 57.47& 28.38& 18.78  & 32.01  & 47.27   & 42.90 & 31.76  & 46.77  & 64.64   \\ \hline
CEGRL-TKGR       & \textbf{42.74}& \textbf{32.32}& \textbf{48.05}& \textbf{62.68}& \textbf{32.90}& \textbf{22.70}& \textbf{36.91}& \textbf{52.95}& \textbf{48.35}& \textbf{37.63}& \textbf{54.22}& 68.47\\ \hline
\end{tabular}}
\end{table*}

\begin{table*}[htb]
\centering
\caption{Experimental results of link prediction on YAGO, WIKI, and GDELT datasets. The best result in each column is boldfaced.}
\label{tab:yagores}\scalebox{0.61}{
\begin{tabular}{ccccccccccccc}
\hline
Model       & \multicolumn{4}{c}{YAGO}                                          & \multicolumn{4}{c}{WIKI}                                          & \multicolumn{4}{c}{GDELT}                                         \\ \hline
            & MRR            & Hits@1         & Hits@3         & Hits@10        & MRR            & Hits@1         & Hits@3         & Hits@10        & MRR            & Hits@1         & Hits@3         & Hits@10        \\
TransE      & 38.97          & 26.87          & 42.45          & 56.05          & 23.46          & 16.53          & 28.45          & 35.71          & -              & -              & -              & -              \\
Distmult    & 44.05          & 39.19          & 49.70          & 59.94          & 27.96          & 18.84          & 32.45          & 39.51          & 8.61           & 3.91           & 8.27           & 17.04          \\
ComplEx     & 44.09          & 39.33          & 49.57          & 59.64          & 27.69          & 18.67          & 31.99          & 38.61          & 9.84           & 5.17           & 9.58           & 18.23          \\
R-GCN       & 20.25          & 11.25          & 24.01          & 37.30          & 13.96          & 7.21           & 15.75          & 22.05          & 12.17          & 8.64           & 12.37          & 20.63          \\ \hline
TTransE     & 31.19          & 18.12          & 40.91          & 51.21          & 29.27          & 21.67          & 34.43          & 42.39          & -              & -              & -              & -              \\
TA-DistMult & 54.92          & 48.15          & 59.61          & 66.71          & 44.53          & 39.92          & 48.73          & 51.71          & 10.34          & 6.25           & 10.44          & 21.63          \\
TNTComplEx  & 57.98          & 52.92          & 61.33          & 66.69          & 45.03          & 40.04          & 49.31          & 52.03          & 19.53          & 12.41          & 20.75          & 33.42          \\
Evo-KG      & 68.81          & 54.49          & 81.40          & \textbf{92.41} & 67.44          & 54.63          & 79.36          & \textbf{85.98} & 18.94          & 11.31          & 20.08          & 34.01          \\
GHT         & 57.22          & 51.64          & 60.68          & 67.17          & 48.50          & 45.08          & 50.87          & 53.69          & 20.04          & 12.68          & 21.37          & 34.42          \\
xERTE       & 84.19          & 80.09          & 88.02          & 89.78          & 73.60          & 69.05          & 78.03          & 79.73          & 19.45          & 11.92          & 20.84          & 34.18          \\
TITer       & 87.47          & 80.09          & \textbf{89.96} & 90.27          & 73.91          & 71.70          & 75.41          & 76.96          & 18.19          & 11.52          & 19.20          & 31.00          \\
RE-GCN      & 82.30          & 78.83          & 84.27          & 88.58          & 78.53          & 74.50          & 81.59          & 84.70          & 19.69          & 12.46          & 20.93          & 33.81          \\
CEN         & 83.49          & 79.66          & 86.10          & 90.04          & 78.52          & 74.65          & 81.44          & 84.59          & 19.96          & 11.39          & 20.97          & 33.77          \\ \hline
CEGRL-TKGR       & \textbf{86.25} & \textbf{82.92} & 88.72          & 91.70          & \textbf{79.66} & \textbf{75.73} & \textbf{82.83} & 85.59          & \textbf{20.11} & \textbf{12.73} & \textbf{21.46} & \textbf{34.51} \\ \hline
\end{tabular}}
\end{table*}

\noindent\textbf{Baselines}. For the link prediction task, we compare CEGRL-TKGR model with two categories of KGR models: (1) \emph{static KGR models}, including TransE~\cite{bordes2013translating}, DistMult~\cite{yang2015embedding}, ComplEx~\cite{trouillon2016complex} and R-GCN\cite{schlichtkrull2018modeling}. We apply these models in static KGs that ignore timestamp information. (2) \emph{TKGR models}, including TTransE~\cite{leblay2018deriving}, TA-DistMult~\cite{garcia2018learning}, TNTComplEx~\cite{lacroix2019tensor}, RE-GCN~\cite{li2021temporal}, GHT~\cite{sun2022graph}, EvoKG~\cite{park2022evokg}, TITer\cite{sun2021timetraveler}, xERTE~\cite{han2020xerte}, TLogic\cite{liu2022tlogic} and CEN\cite{li-etal-2022-complex}. 

\noindent\textbf{Evaluation Metrics}.
The mean reciprocal rank (MRR) and Hits@k are standard metrics for the TKG link prediction task. MRR is the average reciprocal of the correct query answer rank. Hits@$k$ indicates the proportion of correct answers among the top $k$ candidates. We use a more reasonable time-aware filter setting to report our experimental results\footnote{The time-aware filtering setting filters out only the four groups that occur at query time and can simulate extrapolated prediction tasks in the real world~\cite{sun2021timetraveler}.}. 

\noindent\textbf{Implementation Details}. The whole of training hyper-parameters and model configurations are summarized in Appendix A.1. 

\subsection{Experimental Results and Discussion}
Table~\ref{tab:icewsres} and Table~\ref{tab:yagores} report the experimental results of the link prediction task on six benchmark datasets. Static KG embedding methods fell far behind CEGRL-TKGR due to their inability to capture temporal dynamics in the TKG. Our method is also superior to other TKGR models in predicting events. The improved performance shows that surface patterns and noise are widely present in several real-world TKG datasets. The previous models are generally inadequate in design. CEGRL-TKGR based on evolutionary representation will learn the inherent confounding features in the TKG when gathering neighborhood information and transmitting historical information, and the model based on rule-based inference will mine the false correlation in the data, all of which will lead to the model-making non-causal predictions in the reasoning stage. Our model incorporates causal theory into the TKGR task and visibly separates causal features from confounding features. This helps to protect the model from surface patterns and noise present in the dataset and to uncover the real reasons that affect the formation of links between entities. TiTer and EvoKG show excellent performance on YAGO datasets because the former's historical fact search strategy works well on smaller datasets, while the latter's modeling of event timing works well on datasets containing events at relatively regular time intervals. More model configurations and experimental results are summarized in the Appendix.

\section{Conclusion}
\label{sec:conclusion}
In this paper, we revisit the GNN-based TKGR model from the causality perspective, on this basis, we propose a novel CEGRL-TKGR framework. By synergistically integrating causal structures with graph representation learning of the TKG, our framework overcomes the problem of existing TKGR models' learning biased data representations and mining for false correlations unintentionally. Comprehensive experimental results and analysis have proved the effectiveness of CEGRL-TKGR.

\pheadWithSpace{Limitations and Future Work}
The proposed CEGRL-TKGR is an innovational causal enhanced graph representation learning framework for optimizing feature representations directly using causal technology for the TGKR task. The limitations of CEGRL-TKGR are as follows:
\begin{itemize}
    \item From the dataset's perspective, our research primarily focuses on TKG datasets, which may not verify the generalization ability of the CEGRL-TKGR framework to those time-interval insensitive graph datasets. Additionally, we aim to further conduct case studies to enhance the interpretability of the framework in the reasoning procedure as future work.

    \item From the model’s perspective, our research evaluates the TKGR task alone. Theoretically, the GNN-based reasoning paradigm incorporated in the causal structure can be applied to any other graph representation learning tasks, such as triple classification~\cite{jaradeh2021triple}, triple set prediction~\cite{zhang2024start}, and graph classification~\cite{liu2023survey}. In future work, we desire to explore powerful disentanglement methods and more advanced causal intervention strategies to improve the CEGRL-TKGR's performance for more rich graph representation learning-based tasks. Besides, the increased complexity of causal reasoning in the TKG is untouched.

    \item From the adaptation's perspective, to adapt the CEGRL-TKGR framework to more models, there are some hyper-parameters to control causal intervention and training. These hyper-parameters are sensitive to different models and datasets, hence it needs to take sufficient time to experiment to find the optimal values and combinations among them. Therefore, how to reduce the consumption in the above adaptation procedure upon the framework is worthy of consideration.
\end{itemize}

\pheadWithSpace{Acknowledgements} This work was supported by the National Natural Science Foundation of China (Grant No. 62202075, No. 62171111, No. 62376043), the Natural Science Foundation of Chongqing, China (No. CSTB2022NSCQ-MSX1404, No. CSTB2023NSCQ-MSX0091), Fundamental Research Funds for the Central Universities (No. SWU-KR24008), Science Research Project, Chongqing College of International Business and Economics (No. KYZK2024001), Key Laboratory of Data Science and Smart Education, Hainan Normal University, Ministry of Education (No. DSIE202206), and the State Key Laboratory of Public Big Data, Guizhou University (No. PBD2024-0501), Ming Liu's research was funded in part by Australian Research Council Linkage (LP220200746). Yongbin Qin's work was supported by 
the National Natural Science Foundation of China (No. 62166007, No. 62066008) and the National Key R\&D Program of China (No. 2023YFC3304500).

\bibliography{custom}

\appendix

\section{Appendix}
\label{sec:appendix}
\subsection{Implementation Details}
We set the dimension of all embeddings and hidden states to 200. The number of layers of the R-GCN is set to 1 for YAGO and 2 for the other datasets. The optimal number of historical snapshots is set to 8, 10, 10, 1, 2, and 6 for ICEWS14, ICEWS18, ICEWS05-15, YAGO, WIKI, and GDELT, respectively. To fair comparison, static graph constraints are added for ICEWS14, ICEWS18, and ICEWS05-15. The channel number for decoding is set to 50, and the kernel size is set to 4$\times$3. We try several different values for $\lambda_{1},\lambda_{2}$, and $\lambda_{3}$, and finally chose 0.5, 0.5, 0.3. We use Adam to optimize the parameters, with a learning rate of 0.001. All of the experiments are processed on a Linux server with CPU Xeon Gold 6142, RAM 64G, and Nvidia 4090 GPU.

\subsection{Ablation Study}
We investigate the effectiveness of causally enhanced and time-interval guided decoders for the link prediction task. Specifically, CEGRL-TKGR w/o TD means that no time interval vector is used to guide the decoder to work, and CEGRL-TKGR w/o CE means that the model removes causal decoupling and causal intervention parts. Table \ref{tab:ablation} shows the results of ablation experiments, which indicate the effectiveness of these two components. As can be seen from the results in the table, for datasets such as YAGO and WIKI that contain relatively regular time intervals, a temporal gap-guided decoder can capture this time interval pattern well enough to make accurate predictions. At the same time, it does not degrade performance even for time-interval insensitive datasets. Our causal enhancement module, under the independent constraint of emphasizing causal features and confounding features, eliminates the influence of the fast bridge through causal intervention, forcing the model to learn the intrinsic causes of the events. It is worth noting that our causal enhancement module can be seen as a flexible component that can be easily used in several GNN-based reasoning frameworks.

\begin{table*}[ht]
\centering
\caption{The ablation study of our model on the six benchmark datasets. "w/o" means "without". }
\label{tab:ablation}\scalebox{0.63}{
\begin{tabular}{ccccccccccccc}
\hline
Model        & \multicolumn{2}{c}{ICEWS14} & \multicolumn{2}{c}{ICEWS18} & \multicolumn{2}{c}{ICEWS05-15} & \multicolumn{2}{c}{YAGO} & \multicolumn{2}{c}{WIKI} & \multicolumn{2}{c}{GDELT} \\ \hline
             & MRR         & Hits@10       & MRR         & Hits@10       & MRR           & Hits@10        & MRR        & Hits@10     & MRR        & Hits@10     & MRR        & Hits@10      \\
CEGRL-TKGR w/o TD & 42.21       & 62.43         & 32.67       & 52.68         & 48.13         & 68.33          & 84.71      & 90.56       & 78.54      & 84.37       & 19.93      & 34.50        \\
CEGRL-TKGR w/o CE & 41.89       & 61.65         & 32.62       & 52.54         & 48.03         & 68.20          & 81.93      & 88.39       & 79.04      & 84.79       & 19.66      & 33.71        \\
CEGRL-TKGR        & \textbf{42.74}       & \textbf{62.68}         & \textbf{32.90}       & \textbf{52.95}         & \textbf{48.35}         & \textbf{68.47}          & \textbf{86.25}      & \textbf{91.70}       & \textbf{79.66}      & \textbf{85.59}       & \textbf{20.11}      & \textbf{34.51}        \\ \hline
\end{tabular}}
\end{table*}

\subsection{Parameter Sensitivity Analysis}
\begin{figure*}[t]
\centering
    \subfigure{
\includegraphics[width=0.45\textwidth]{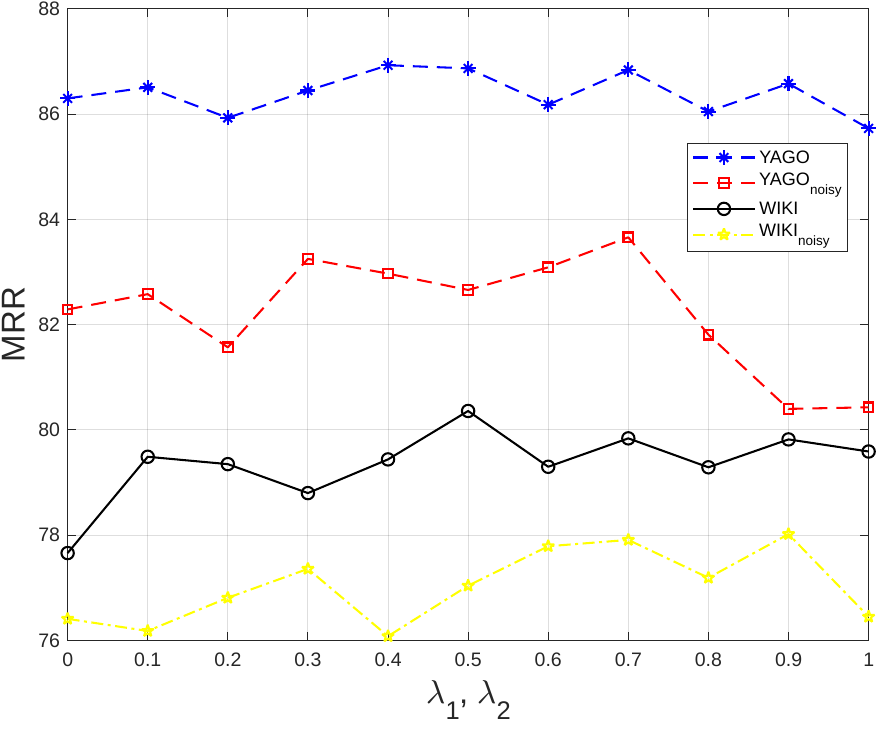}}
    \subfigure{    \includegraphics[width=0.45\textwidth]{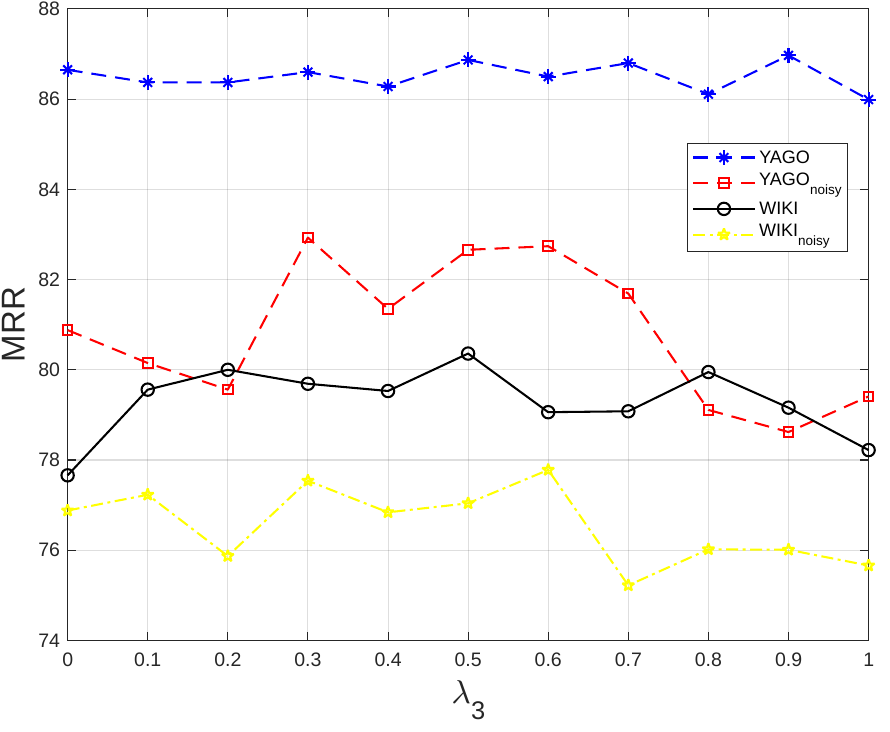}}
    \vspace{-5pt}
    \caption{The parameters sensitivity analysis of loss coefficients $\lambda_{1}, \lambda_{2}$ and $\lambda_{3}$.} 
\label{fig:hyper}
\end{figure*}

In the CEGRL-TKGR, $\lambda_{1}$ and $\lambda_{2}$ jointly affect the disentanglement intensity of causal and confounding features, and $\lambda_{3}$ controls the intensity of causal intervention. We study the sensitivity of parameters in different benchmark datasets, as depicted in Fig.~\ref{fig:hyper}. Specifically, one parameter is fixed at 0.5 and the other parameter varies in [0,1] with a step size of 0.1. The model is relatively stable in most parameter selection cases, but on noisy datasets, the model has higher requirements for hyper-parameters, and extreme values will degrade the performance of the model. The best range for $\lambda_{1}$, $\lambda_{2}$ is about 0.5 to 0.7. $\lambda_{3}$ should be a relatively small value, ranging from 0.3 to 0.6. 

\begin{figure*}[ht]
\centering
    \subfigure[MRR results on the YAGO dataset.]{
    \includegraphics[width=0.44\textwidth]{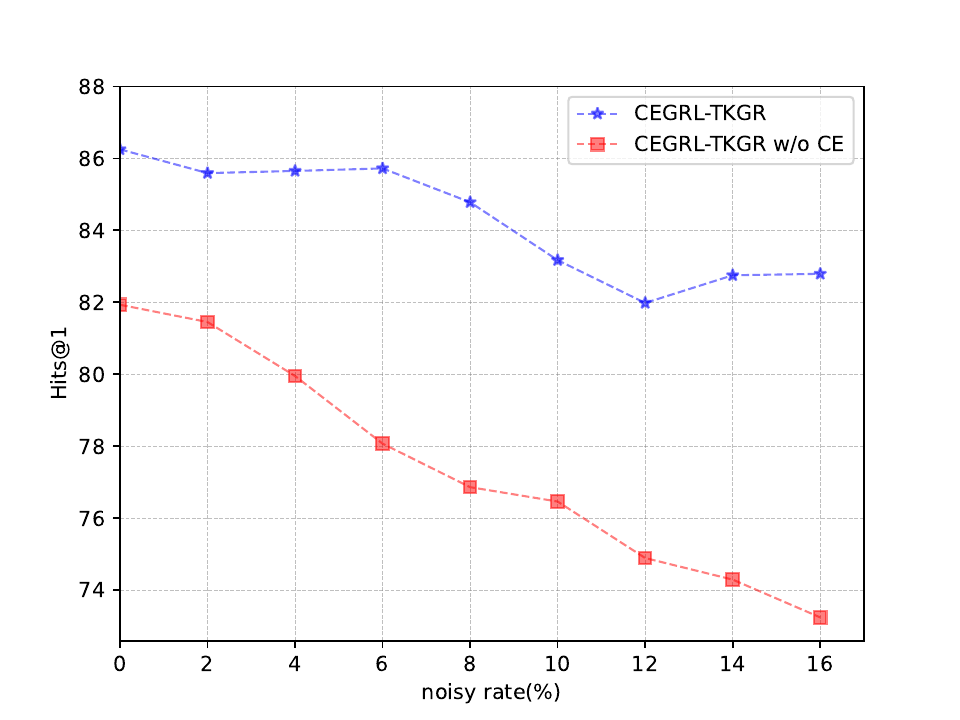}
    }
    \hspace{0.2in}
    \subfigure[Hits@1 results on the YAGO dataset.]{
    \includegraphics[width=0.44\textwidth]{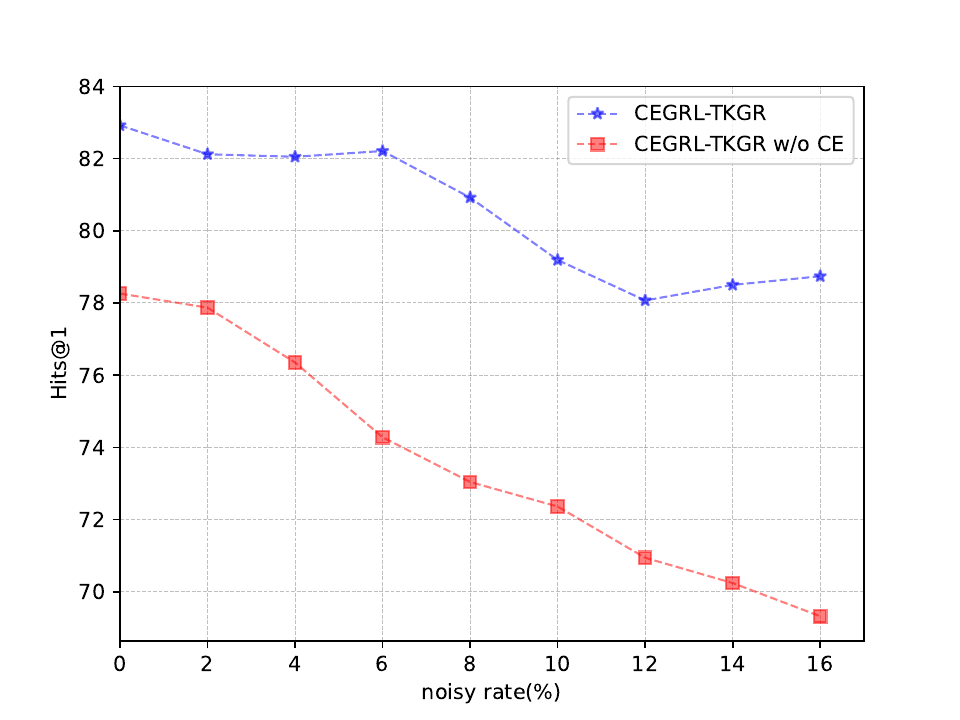}
    }
    \subfigure[MRR results on the WIKI dataset.]{
    \includegraphics[width=0.44\textwidth]{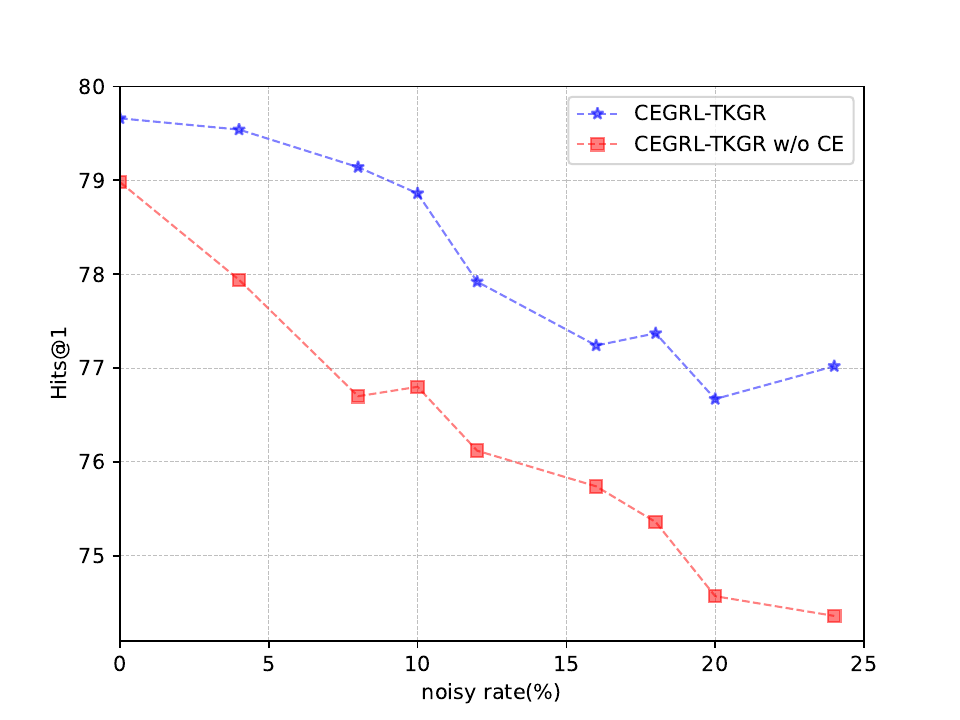}
    }
    \hspace{0.2in}
    \subfigure[Hits@1 results on the WIKI dataset.]{
    \includegraphics[width=0.44\textwidth]{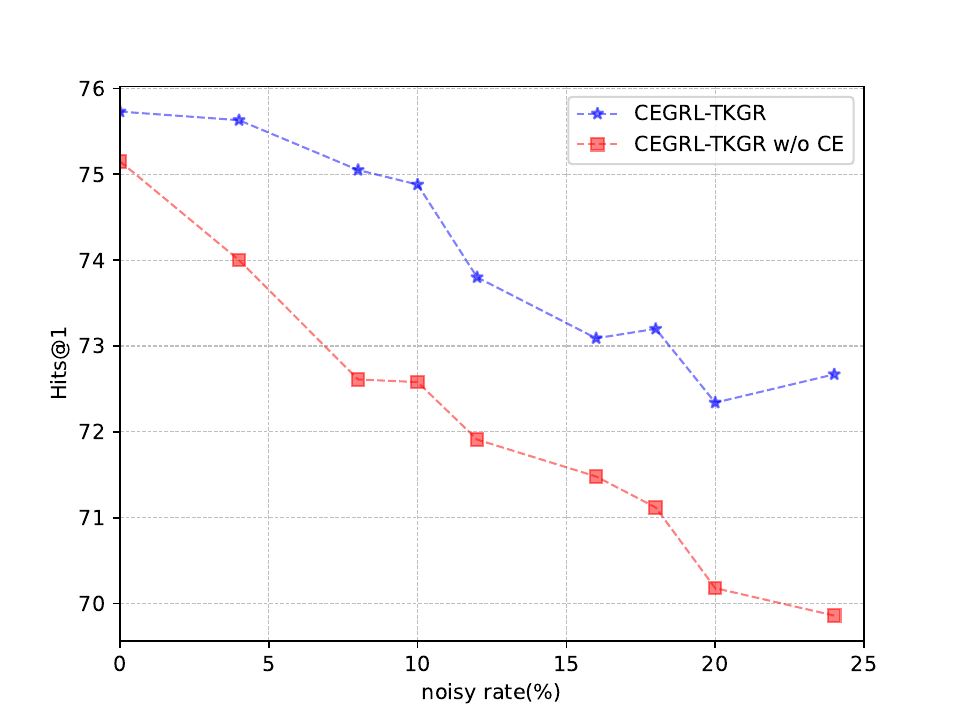}
    }
    \caption{The performance of  CEGRL-TKGR and CEGRL-TKGR w/o CE on the noisy YAGO and WIKI datasets, respectively.}
\label{fig:noise}
\end{figure*}

\subsection{Performance on Noisy Temporal Knowledge Graphs}
To explore whether the proposed CEGRL-TKGR can alleviate noise and surface patterns, we randomly replace a certain percentage of positive triples in the training set of each TKG dataset in form of noisy TKGs. Taking YAGO and WIKI datasets as examples, we test the performance of CEGRL-TKGR and CEGRL-TKGR w/o CE under different noise deviations, respectively. The experimental results are shown in Fig.~\ref{fig:noise}.

From the experimental results, we can draw the following conclusion: when the noise in the dataset increases, the performance of models lacking the recognition of causal features and confounding features will deteriorate sharply, and the performance of MRR and Hits@1 will decrease, which indicates that the CEGRL-TKGR w/o CE is easy to capture data bias and make wrong predictions based on it. In contrast, CEGRL-TKGR uses the causal enhancement module to effectively reduce the impact of confounding features and shows more stable performance on these two noisy TKG datasets. The performance degradations on MRR and Hits@1 are significantly smaller than those without the causal module.

\end{document}